\documentclass[10pt,twocolumn,letterpaper]{article}

\usepackage[pagenumbers]{cvpr} 

\usepackage[accsupp]{axessibility}
\usepackage{float}
\usepackage{multirow}
\usepackage{booktabs}
\usepackage{enumitem}
\usepackage{makecell} 
\usepackage{multirow} 
\usepackage[table]{xcolor} 
\usepackage{marvosym}
\usepackage[ruled,vlined,linesnumbered]{algorithm2e}
\usepackage{algpseudocode}
\usepackage{array}
\usepackage[most]{tcolorbox}

\newcommand{\Our}{FlexiMMT\xspace}
\newcommand{\myparagraph}[1]{{\setlength{\parskip}{0.3em} \noindent \textbf {#1}}}

\definecolor{myyellow}{RGB}{254, 245, 225}
\definecolor{mycolor1}{rgb}{0.85, 0.9, 1} 
\definecolor{mycolor2}{rgb}{0.98,0.86,0.87} 
\definecolor{darkgreen}{rgb}{0,0.69,0.31} 
\definecolor{darkblue}{rgb}{0,0.44,0.753} 

\definecolor{motionred}{RGB}{255, 0, 0}
\definecolor{motionorange}{RGB}{255, 192, 0}
\definecolor{motionyellow}{RGB}{250, 244, 0}
\definecolor{motiongreen}{RGB}{0, 176, 80}
\definecolor{motionblue}{RGB}{0, 176, 240}
\definecolor{motiondarkblue}{RGB}{0, 112, 192}
\definecolor{motionpurple}{RGB}{112, 48, 160}

\newlength{\mycellwidth}
\newcommand{\mystrut}{\vphantom{\mathcal{M}_{y\rightarrow y}}}

\newcommand{\mycellbox}[2]{%
  \colorbox{#1}{\makebox[\mycellwidth][c]{$\mystrut #2$}}%
}

\definecolor{cvprblue}{rgb}{0.21,0.49,0.74}
\usepackage[pagebackref,breaklinks,colorlinks,allcolors=cvprblue]{hyperref}

\def\confName{CVPR}
\def\confYear{2026}

\title{Let Your Image Move with Your Motion! -- Implicit Multi-Object Multi-Motion Transfer}

\author{
Yuze Li\textsuperscript{\rm 1}\quad
Dong Gong\textsuperscript{\rm 2}\quad
Xiao Cao\textsuperscript{\rm 3}\quad
Junchao Yuan\textsuperscript{\rm 1}\quad
Dongsheng Li\textsuperscript{\rm 1}\quad
Lei Zhou\textsuperscript{\rm 4}\\[0.2em]
Yun Sing Koh\textsuperscript{\rm 5}\quad
Cheng Yan\textsuperscript{\rm 1}$^\dagger$\quad
Xinyu Zhang\textsuperscript{\rm 5}$^\spadesuit$\\[0.2em]
\small
\textsuperscript{\rm 1}Tianjin University\quad
\textsuperscript{\rm 2}University of New South Wales\quad
\textsuperscript{\rm 3}University of Electronic Science and Technology of China\quad
\textsuperscript{\rm 4}Hainan University\\
\small
\textsuperscript{\rm 5}University of Auckland\\[0.5em]
}

\begin{document}
\twocolumn[{%
\renewcommand\twocolumn[1][]{#1}%
\maketitle
\begin{center}
    \centering
    \captionsetup{type=figure}
    \vspace{-1cm}
    \includegraphics[width=\textwidth]{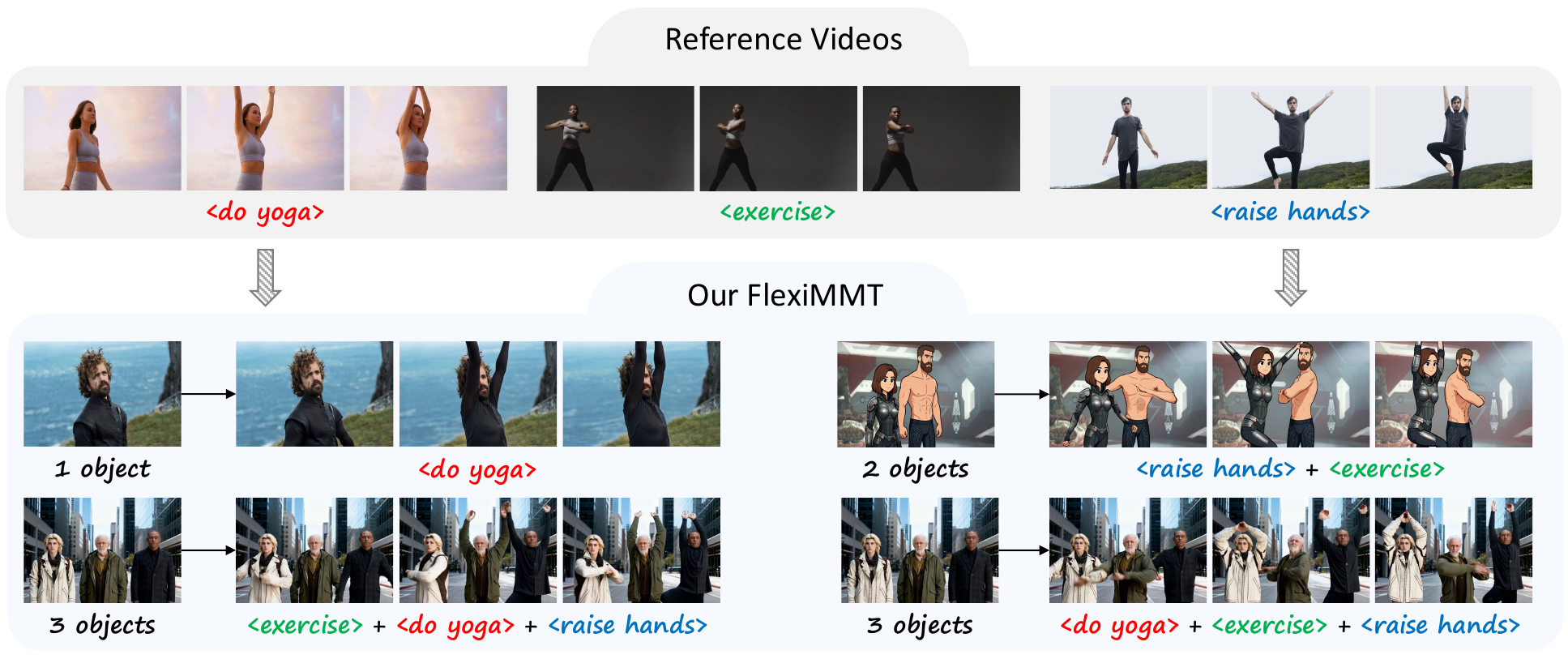}
    \captionof{figure}{\textbf{Illustration of \Our.} 
    Given multiple reference videos, our \Our independently extracts each motion and applies them to images with any number of objects, enabling precise and compositional multi-object multi-motion transfer. For clarity, we show only simplified motion labels here; full text prompts and vivid videos are provided in the Supplementary Material.}
    \label{fig:big_image}
\end{center}%
}]

\maketitle

\begingroup
\renewcommand\thefootnote{}
\footnotetext{$^\spadesuit$ Project Lead.}
\footnotetext{$^\dagger$ Corresponding Author}
\addtocounter{footnote}{-2}
\endgroup

\begin{abstract}
Motion transfer has emerged as a promising direction for controllable video generation, yet existing methods largely focus on single-object scenarios and struggle when multiple objects require distinct motion patterns. In this work, we present \Our, the first implicit image-to-video (I2V) motion transfer framework that explicitly enables multi-object, multi-motion transfer. Given a static multi-object image and multiple reference videos, \Our independently extracts motion representations and accurately assigns them to different objects, supporting flexible recombination and arbitrary motion-to-object mappings. To address the core challenge of cross-object motion entanglement, we introduce a Motion Decoupled Mask Attention Mechanism that uses object-specific masks to constrain attention, ensuring that motion and text tokens only influence their designated regions. We further propose a Differentiated Mask Propagation Mechanism that derives object-specific masks directly from diffusion attention and progressively propagates them across frames efficiently. Extensive experiments demonstrate that \Our achieves precise, compositional, and state-of-the-art performance in I2V-based multi-object multi-motion transfer. Our project page is: \small\url{https://ethan-li123.github.io/FlexiMMT_page/}

\end{abstract}
\section{Introduction}
\label{sec:intro}

Recent advancements in video diffusion models~\cite{animatediff, svd, cogvideox, wan, dynamicrafter, liao2024evaluation} have significantly accelerated the development of controllable video generation~\cite{motiondirector, vmc, dmt, motionclone, videomage, loraedit, go-with-the-flow, keyframe, motionpro, vace, animateanyone, motionprompting}. 
Among various controllable tasks, motion transfer has emerged as a key direction, aiming to capture motion dynamics from reference guidances and apply them to target subjects.
Some motion transfer methods use explicit motion guidance, such as pose \cite{animateanyone, wananimate, followyourpose, disco, magicanimate}, optical flow~\cite{go-with-the-flow, mofavideo, motioni2v}, or trajectories \cite{peekaboo, revideo, motionprompting, draganything, motionpro, zhang2025s2v2v}.
While effective, these approaches rely on additional estimators and require accurate motion extraction, which can be computationally expensive.
In contrast, implicit-motion methods encode motion directly from reference videos into latent embeddings or learned parameters~\cite{anyv2v, flexiact, motiondirector, vmc, dmt, motionclone, dualreal, amf, videopainter, liu2025multimotion, ma2025follow}. This paradigm avoids explicit motion preprocessing and provides greater flexibility and ease of use.
In this paper, we focus on implicit motion transfer from reference videos.

Based on the type of input conditions,
existing motion transfer methods can be grouped into two major directions, \ie, text-to-video (T2V)~\cite{motiondirector, vmc, dmt, motionclone, videomage, loraedit} and image-to-video (I2V)~\cite{animateanyone, go-with-the-flow, motionprompting}. T2V methods typically transfer motion from reference sequences while synthesizing semantic content guided by text descriptions.
I2V methods are more challenging as they must simultaneously preserve the appearance of a given first-frame image and follow the reference motion, requiring much stronger spatial and structural consistency.
Most recent studies~\cite{anyv2v, i2vedit, flexiact, loraedit} focus on single-concept (\ie, one object or one motion) motion transfer. 
However, \textit{simultaneously transferring distinct motions to multiple objects within the same image remains largely unexplored} and continues to be an open challenge.

In this paper, we propose \textbf{\Our}, a novel implicit motion transfer framework that, to the best of our knowledge, is the \textbf{first} to explicitly tackle \textbf{multi-object, multi-motion} challenges in I2V generation.
Given a multi-object input image and multiple reference videos, our method independently extracts each motion and accurately assigns it to the corresponding target object, supporting flexible, compositional, and arbitrary motion–object rearrangements.

The central challenge in this setting is the motion entanglement across different objects.
To this end, we introduce a \textit{Motion Decoupled Mask Attention Mechanism} (MDMA), which leverages object-specific masks to perform explicit motion disentanglement within the attention layers. 
This mechanism enforces that text and motion tokens exclusively attend to their corresponding objects from video tokens rather than interfering with others, thereby enabling precise and independent motion control.
To efficiently obtain object-specific masks, we introduce a \textit{Differentiated Mask Extraction Mechanism} (DMEM). Unlike approaches that rely on external segmentation models~\cite{sam,sam2,groundedsam}, 
DMEM directly derives object masks from the model’s own attention activations, providing an efficient training-time solution.
At inference, we further mitigate cross-object mask confusion through a dynamic \textit{Regressive Mask Propagation Mechanism} (RMPM) that progressively and efficiently transfers first-frame masks to subsequent frames, maintaining clean and consistent object boundaries throughout the generation process.

Quantitative and qualitative experiments on 200 video-image pairs demonstrate that our \Our enables independent and flexible motion assignment across multiple objects. As illustrated in \cref{fig:big_image}, 
different motion patterns extracted from multiple reference videos can be accurately transferred to distinct objects in the input image, and these motions can be arbitrarily swapped or recombined across objects.
Our contributions are as follows:
\begin{itemize}
    \item We introduce \Our, the first implicit-motion I2V framework capable of multi-object, multi-motion transfer. 
    \item We design a Motion Decoupled Mask Attention Mechanism that uses object-specific masks to disentangle motion across objects, enabling precise and flexible per-object motion assignment.
    \item We present a Differentiated Mask Extraction Mechanism that derives object-specific masks directly from diffusion attention with dynamic progressive propagation scheme for stable multi-object control.
    \item Extensive experiments demonstrate that \Our offers strong flexibility, compositionality, and state-of-the-art performance in multi-object I2V motion transfer.
\end{itemize}
\section{Related Work}
\label{sec:related}

\begin{figure*}[t!]
    \centering
    \includegraphics[width=\linewidth]{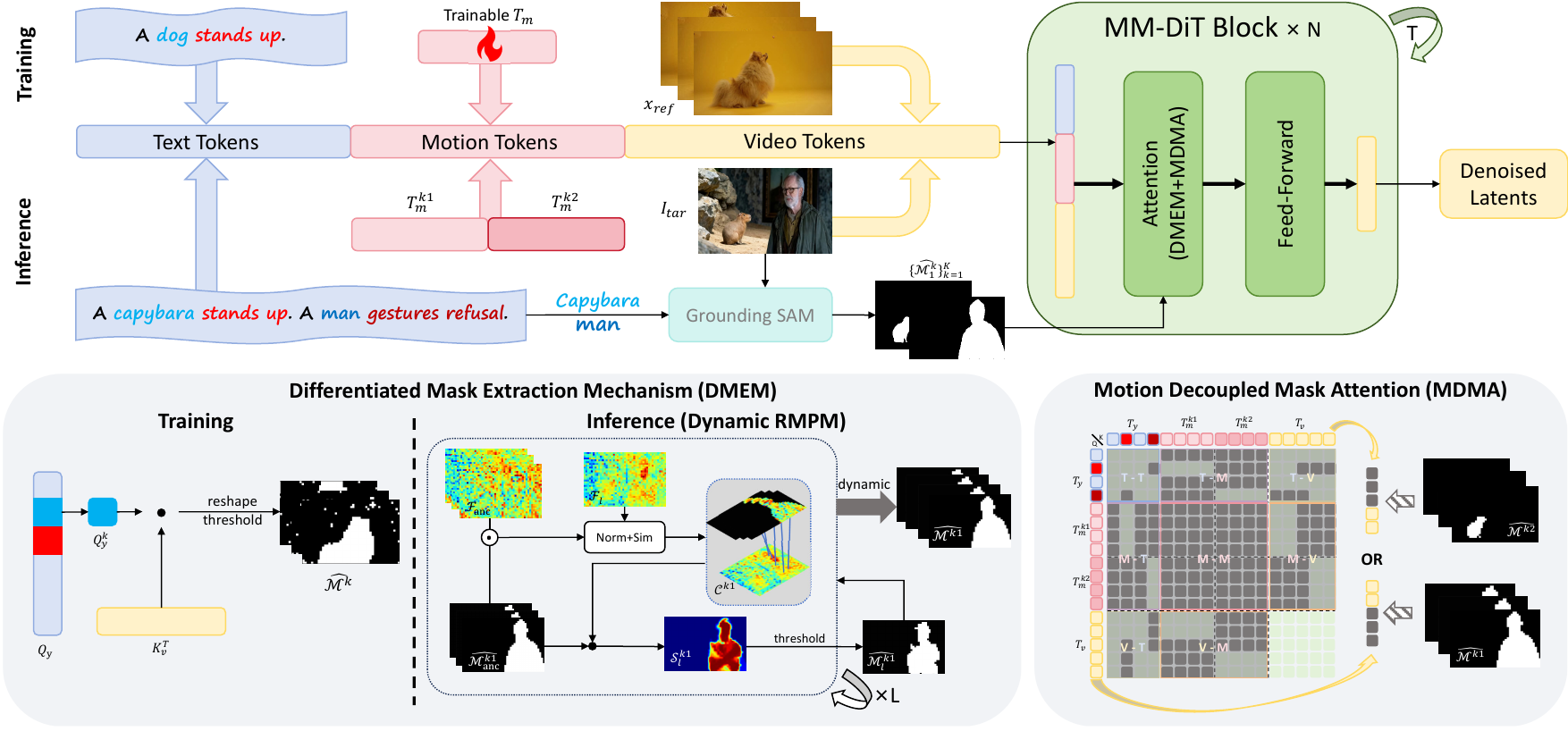}
    \caption{\textbf{Overview of FlexiMMT.} (a) Training: Given one reference video, insert trainable motion tokens into text and video tokens. Get the object mask through a simple $QK$ multiplication method, then mask out M2M, M2V and T2M parts in attention map. (b) Inference: Given a multi-object conditional image, we first segment each object's mask with semantic segmentation model \cite{groundedsam}. Concatenate pre-trained motion tokens into text and video tokens for inference. Extract each object's latent-space mask in subsequent frames via Dynamic Regressive Mask Propagation Mechanism (Dynamic RMPM), and apply it to Motion parts and Text parts in attention map. }
    \label{fig:train_inf}
    \vspace{-0.3cm}
\end{figure*}

\subsection{Explicit Motion Representation Models}
Explicit motion representation models achieve controllable video generation through predefined motion signals such as pose skeletons, optical flow fields, and trajectories. The core idea is to enforce explicit constraints to ensure that the generated motion aligns with the desired behavior.

In the T2V domain, methods are categorized by motion signals: pose-driven method like FollowYourPose~\cite{followyourpose} focus on fusing text and pose sequences to address the limitation of text in specifying fine-grained human motion; depth/edge-based methods like ControlVideo~\cite{controlvideo} and SparseCtrl~\cite{sparsectrl} adopt ControlNet-like~\cite{controlnet} structures to condition generation on text and depth/edge signals for temporal coherence; trajectory-based method like Peekaboo~\cite{peekaboo} use interactive frameworks with text and foreground trajectories to guide target motion along predefined paths.

In the I2V domain, models emphasize appearance-motion synergy: pose-driven methods such as AnimateAnyone~\cite{animateanyone} and MagicAnimate~\cite{magicanimate} fuse pose and reference image features, with enhancements to temporal consistency; optical-flow-driven methods like Go-with-the-Flow~\cite{go-with-the-flow} and MotionI2V~\cite{motioni2v} leverage optical flow for noise warping and spatial alignment with appearance; trajectory-driven methods such as DragAnything~\cite{draganything}, S$^2$V2V~\cite{zhang2025s2v2v}, ReVideo~\cite{revideo}, and MotionPro~\cite{motionpro} use predefined trajectories or trajectory-matching losses to control motion from reference images; other structure-signal-driven methods like Keyframe~\cite{keyframe}, VideoRepainter~\cite{videopainter}, MotionBooth~\cite{motionbooth}, and VACE~\cite{vace} rely on edited keyframes, pose/trajectory integration, or multi-task control signals for guided generation.
Some works~\cite{ruan2023mm, zhang2025let,alexanderson2023listen} leverage multimodal signals to guide and control video motion.

Explicit models require manual annotations or complex preprocessing and are often restricted by the geometry of the source objects. In contrast, our I2V-based method automatically extracts implicit motion from reference videos, eliminating the need for manual labeling and enabling flexible motion transfer across diverse objects and scenes.

\subsection{Implicit Motion Representation Models}
Implicit motion representation models extract motion-related vectors, attention maps, or parameters from reference videos to achieve motion capture and transfer.

In the T2V domain, two categories prevail: direct motion feature extraction methods such as VMC~\cite{vmc}, DMT~\cite{dmt}, and AMF~\cite{amf} capture dynamic cues, relative motion variations, or patch-level flow via temporal attention or intermediate features; LoRA-based methods like MotionDirector~\cite{motiondirector}, VideoMage~\cite{videomage}, and DIVE~\cite{dive} disentangle appearance and motion using spatiotemporal LoRA, with adaptations for concept-specific transfer.

In the I2V domain, models focus on decoupling appearance and motion: feature-based methods such as AnyV2V~\cite{anyv2v} and FlexiAct~\cite{flexiact} construct guidance signals from reference videos' spatiotemporal attention or frequency features; LoRA-based methods like I2VEdit~\cite{i2vedit} and LoRA-Edit~\cite{loraedit} train motion LoRA modules on temporal attention layers to enhance long-term consistency.

However, most implicit methods remain limited to single-object or fixed-layout motion transfer. Our approach overcomes these constraints by extracting implicit representations that scale to multiple objects and generalize across diverse spatial configurations.
\section{Method}
\label{sec:method}

\subsection{Preliminary}
\label{Preliminary}
\myparagraph{Problem definition.} The multi-object, multi-motion I2V task aims to transfer distinct motion patterns to a given multi-object image, such that each object in the image ``moves'' according to its corresponding reference motion.
Let $\{x_\mathrm{ref}^i\}_{i=1}^N$ denote a set of reference videos, each with corresponding caption $\{y_\mathrm{ref}^i\}_{i=1}^N$, where $N$ is the total number of these videos.
Given a user-specified target image $I_\mathrm{tar}$ containing multiple objects $\{o_\mathrm{tar}^k\}_{k=1}^K$, where $K$ denotes the number of objects, our objective is to generate a target video $x_\mathrm{tar}$.
The generated video should preserve the appearance of the input image in its first frame, while ensuring that each object in $I_\mathrm{tar}$ follows the motion pattern extracted from its designated reference video.

\myparagraph{Image-to-Video (I2V) diffusion model.}
I2V diffusion models generate videos by iteratively denoising a sequence of Gaussian-noise latents, conditioned on a first-frame image and a text prompt.
They are commonly parameterized as $\epsilon_\theta(\mathbf{z}_t, t, c, \mathbf{I}_0)$, using a 3D U-Net or a Transformer. At each diffusion step $t$, the model predicts the noise added to a noisy latent $\mathbf{z}_t$, conditioned on text prompt $y$ and the latent of first-frame image $\mathbf{I}_0$. The training objective follows the standard noise-prediction loss:
\begin{equation}
\mathcal{L} = \mathbb{E}_{y, \mathbf{I}_0, \epsilon, t} \left[ \left\| \epsilon_\theta(\mathbf{z}_t, y, t, \mathbf{I}_0) - \epsilon \right\|_2^2 \right].
\end{equation}

\myparagraph{Motion-based DiT methods.}
For video diffusion models~\cite{cogvideox} using the MM-DiT \cite{dit} architecture, video tokens $T_v$ and text tokens $T_y$ interact within the 3D full-attention layers. Similar to recent works~\cite{flexiact, vfxmaster, odiscoedit}, we introduce trainable tokens called motion tokens $T_m$ to model motion-specific features. After linear projection, the model produces separate query, key, and value representations for text, video, and motion tokens, denoted as:
$(Q_{y\;}, K_{y\;}, V_{y\;}); (Q_{v\;}, K_{v\;}, V_{v\;}); (Q_{m\;}, K_{m\;}, V_{m\;})$.
The attention computation is thus formulated as:
\begin{equation}
\centering
\begin{split}
\mathcal{A} = \frac{[Q_c \oplus Q_m \oplus Q_v][K_c \oplus K_m \oplus K_v]^\top}{\sqrt{d}},
\end{split}
\end{equation}
where $\oplus$ denotes token-wise concatenation.
$\mathcal{A}$ is the attention map. 
Although motion tokens learned from reference videos can be injected into the 3D full-attention layers during inference to achieve motion transfer, this naive strategy cannot assign different motions to different objects within the same image, as the interactions among tokens remain globally entangled.

\subsection{
Motion Decoupled Mask Attention
}
\label{sec:Mask-Attention Motion Decoupling Mechanism}
To address the issue of cross-object motion entanglement, we introduce the \textit{Motion Decoupled Mask Attention} (MDMA) mechanism, enforcing that motion and text tokens interact only with the video tokens belonging to their designated object regions.
Specially, MDMA constructs an object-specific mask matrix $\mathcal{M}$, activating only the token pairs relevant to the target object. This mask is applied to the raw attention map via $\mathcal{A}\leftarrow\mathcal{A} \odot \mathcal{M}$, where $\odot$ denotes element-wise multiplication. By suppressing irrelevant cross-object interactions, MDMA explicitly disentangles motion at the attention level.

Given a target object $o_\mathrm{tar}^{k}$, the $k$-th object in $I_\mathrm{tar}$, we pre-defined notations to clearly distinguish its associated tokens. Specifically, we denote the text, motion, and video tokens for the $k$-th object $o_\mathrm{tar}^{k}$ as $T_y^k$, $T_m^k$, and $T_v^k$, respectively. Since only a subset of text tokens encodes actual motion semantics, we further denote $T^{k}_{y,mo}$ as text tokens that correspond to the motion description of the $k$-th object.
To obtain the video tokens corresponding to object $k$: $T_v^k$, a spatial mask $\widehat{\mathcal{M}}^k$ is required to isolate the video tokens of object $o_\mathrm{tar}^{k}$.
Details of constructing $\widehat{\mathcal{M}}^k$ are provided in Section~\ref{sec:Differentiated Mask Extraction Mechanism}.
For clarity, we illustrate the formulation using two objects, \ie, $o_\mathrm{tar}^{k1}$ and $o_\mathrm{tar}^{k2}$, which can be easily extended to multiple objects.
Here, $K=2$.

The object-specific mask matrix $\mathcal{M}$ in our MDMA is built with two main parts, including Motion-to-[X] (M2X) mask and Text-to-[X] (T2X) mask.

\myparagraph{Motion-to-[X] (M2X) mask.}
M2X aims to make motion-related tokens only focus on the specified object, including Motion-to-Video, Video-to-Motion, and Motion-to-Motion. 

For Motion-to-Video mask, we generate $\mathcal{M}_{m\rightarrow v}$ where if and only if both motion and video tokens belong to this object. Given the $k$-th object, this mask is formulated as:
\begin{equation}
\begin{split}
\mathcal{M}_{m\rightarrow v}^k[p,q] &=
\begin{cases}
1, & \text{if}~~ T_m[p] \in T_m^k \ \land \ T_v[q] \in T_v^k \\
0, & \text{otherwise}
\end{cases}
\end{split}
,
\end{equation}
where $p$ and $q$ are the relative position in motion and video tokens, respectively.
If we have two objects, the $\mathcal{M}_{m\rightarrow v}$ is the union of each mask, represented as:
\begin{equation}
\begin{split}
\mathcal{M}_{m\rightarrow v}[p,q] &= \mathcal{M}^{k1}_{m\rightarrow v}[p,q] \cup \mathcal{M}^{k2}_{m\rightarrow v}[p,q],
\end{split}
\end{equation}
based on the symmetry propority, Video-to-Motion mask is the transpose of Motion-to-Video mask, \ie, $\mathcal{M}_{v\rightarrow m}[q,p] = \mathcal{M}_{m\rightarrow v}[p,q]$. 

To prevent motion tokens from affecting each other, we enfore the Motion-to-Motion Mask ($\mathcal{M}_{m\rightarrow m}$) to be a zero matrix as: 
\begin{equation}
\mathcal{M}_{m\rightarrow m} = \mathbf{0}_{(K \cdot d_m) \times (K \cdot d_m)},
\end{equation}
where $d_m$ is the length of the motion tokens $T_m^k$.

\myparagraph{Text-to-[X] (T2X) mask.}
T2X ensures that motion-related text tokens attend only to the video tokens of their corresponding object, preventing cross-object motion interference.
It contains Text-to-Video, Video-to-Text, Text-to-Text, Text-to-Motion and Motion-to-Text.

For Text-to-Video mask $\mathcal{M}_{y\rightarrow v}$, similar as Motion-to-Video mask,
it enforces that only text tokens describing the motion of object $k$ can attend to video tokens belonging to that same object. Formally, for the $k$-th object:
\begin{equation}
\begin{split}
\mathcal{M}_{y\rightarrow v}^k[p,q] &=
\begin{cases}
1, & \text{if}~~T_y[p] \in T_{y,mo}^k \ \land \ T_v[q] \in T_v^k \\
0, & \text{otherwise}
\end{cases}
\end{split}
,
\end{equation}
where $p$ and $q$ here denote relative positions in text and video tokens, respectively.
For two objects, the overall mask $\mathcal{M}_{y\rightarrow v}$ is obtained by merging the per-object masks:
\begin{equation}
\begin{split}
\mathcal{M}_{y\rightarrow v}[p,q] &= \mathcal{M}^{k1}_{y\rightarrow v}[p,q] \cup \mathcal{M}^{k2}_{y\rightarrow v}[p,q],
\end{split}
\end{equation}
the corresponding Video-to-Text mask is symmetric: $\mathcal{M}_{v\rightarrow y}[q,p] = \mathcal{M}_{y\rightarrow v}[p,q]$.

Text-to-Text mask $\mathcal{M}_{y\rightarrow y}$ prevents motion-related text tokens of different objects from attending to each other.
For two objects, the formulation is defined as: 
\begin{equation}
\begin{split}
\mathcal{M}_{y\rightarrow y}[p,q] =
\begin{cases}
0, & \text{if}~~ T_y[p] \in T_{y,mo}^{k1} \ \land \ T_y[q] \in T_{y,mo}^{k2} \\
1, & \text{otherwise}
\end{cases}
\end{split}
.
\end{equation}

Text-to-Motion Mask $\mathcal{M}_{y\rightarrow m}$ is designed to ensure that each text tokens interacts only with the motion tokens of its corresponding object:
\begin{equation}
\begin{split}
\mathcal{M}^k_{y \rightarrow m}[p,q] &=
\begin{cases}
1, & \text{if}~~ T_y[p] \in T_{y,mo}^{k} \ \land \ T_m[q] \in T_m^k \\
0, & \text{otherwise}
\end{cases}
\end{split}
,
\end{equation}
for two objects, the final mask is obtained by taking the union of object-specific masks:
\begin{equation}
\begin{split}
\mathcal{M}_{y\rightarrow m}[p,q] &= \mathcal{M}^{k1}_{y\rightarrow m}[p,q] \cup \mathcal{M}^{k2}_{y\rightarrow m}[p,q],
\end{split}
\end{equation}
the symmetric Motion-to-Text mask is thus defined as $M_{m \rightarrow y}[q,p] = M_{y \rightarrow m}[p,q]$. 

\begin{figure}[t]
    \centering
    \includegraphics[width=0.9\linewidth]{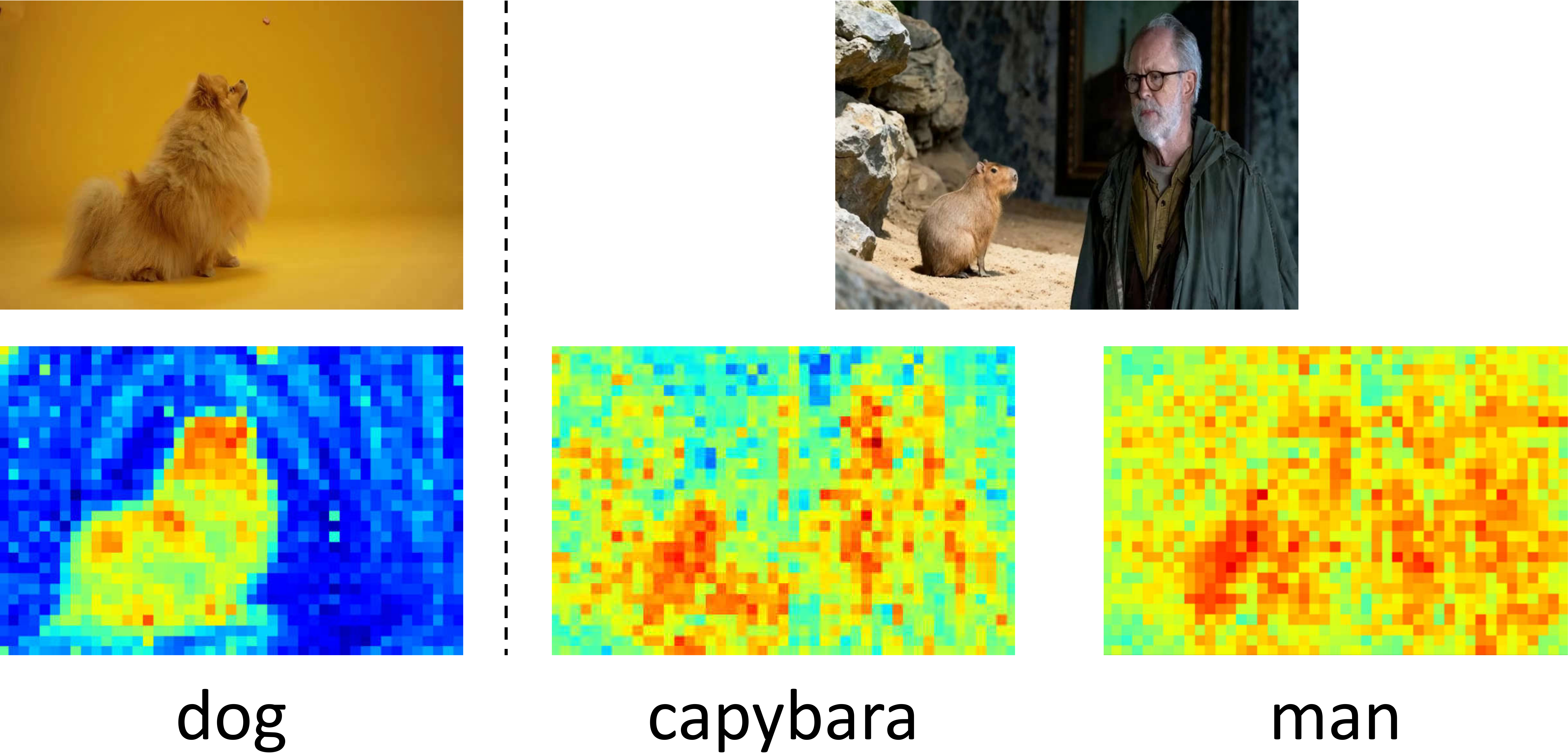}
    \caption{\textbf{Overview of the simple method for training and inference.} Single object can be extracted through a simple $QK$ multiplication method. However, different objects cannot be extracted through this simple method due to feature entanglement.}
    \label{fig:simple_extraction}
    \vspace{-0.3cm}
\end{figure}
\myparagraph{Overall.} The object-specific attention mask $\mathcal{M}$ is constructed by combining the \colorbox{mycolor2}{M2X} and \colorbox{mycolor1}{T2X} sub-masks as:
\begin{equation}
\mathcal{M} = \begin{bmatrix}
\mycellbox{mycolor1}{\mathcal{M}_{y\rightarrow y}} & \mycellbox{mycolor1}{\mathcal{M}_{y\rightarrow m}} & \mycellbox{mycolor1}{\mathcal{M}_{y\rightarrow v}} \\
\mycellbox{mycolor1}{\mathcal{M}_{m\rightarrow y}} & \mycellbox{mycolor2}{\mathcal{M}_{m\rightarrow m}} & \mycellbox{mycolor2}{\mathcal{M}_{m\rightarrow v}} \\
\mycellbox{mycolor1}{\mathcal{M}_{v\rightarrow y}} & \mycellbox{mycolor2}{\mathcal{M}_{v\rightarrow m}} & {\mathbf{I}_{d_v \times d_v}}
\end{bmatrix}
,
\end{equation}
where $\mathbf{I}$ is the identity matrix, and $d_v$ is the number of video tokens $T_v$.
The mask $\mathcal{M}$ is applied to attention map during both inference and training. Since training video contains only a single object, we only active M2M, M2V and T2M sub-masks, while others blocks leave the corresponding attention values unchanged.

\subsection{Differentiated Mask Extraction Mechanism}
\label{sec:Differentiated Mask Extraction Mechanism}
Recall that for the $k$-th object, the object-specific video tokens $T_v^k$ require a spatial mask $\widehat{\mathcal{M}}^k$ to isolate the corresponding object region.
To obtain such object-specific masks $\widehat{\mathcal{M}}=\{\widehat{\mathcal{M}}^k\}_{k=1}^{K}$ for all objects, we propose a \textit{Differentiated Mask Extraction Mechanism} (DMEM), which employs two complementary mask extraction strategies for the training and inference stages.

\myparagraph{Training stage.}
During training, each video contains only one single object; therefore, we set the object index to $k=1$.
Specifically, we first localize the object region and extract its object-specific text query representation $Q_y^k$.
Using $Q_y^k$, we compute an attention map by multiplying $Q_y^k$ and $K_v^\top$, where larger values indicate a higher likelihood that a video token corresponds to the target object.
We then obtain the mask $\widehat{\mathcal{M}}^k$ by applying a hard threshold to this attention map: values above the threshold are set to $1$, and the rest to $0$. For simplicity, we use the mean value of the attention map as the threshold.

\noindent \textit{Discussion.} An alternative way to obtain $\widehat{\mathcal{M}}$ is to use semantic segmentation methods such as SAM~\cite{sam,sam2}. However, we adopt our lightweight attention-based strategy for two reasons:
i) segmentation introduces substantial computational training overhead, while DiT attention already provides reliable object cues;
ii) applying segmentation only during training leads to a train-test inconsistency, as it cannot be used at inference time, potentially harming performance. 
We leave more explorations for future work.

\begin{figure}[t]
    \centering
    \includegraphics[width=\linewidth]{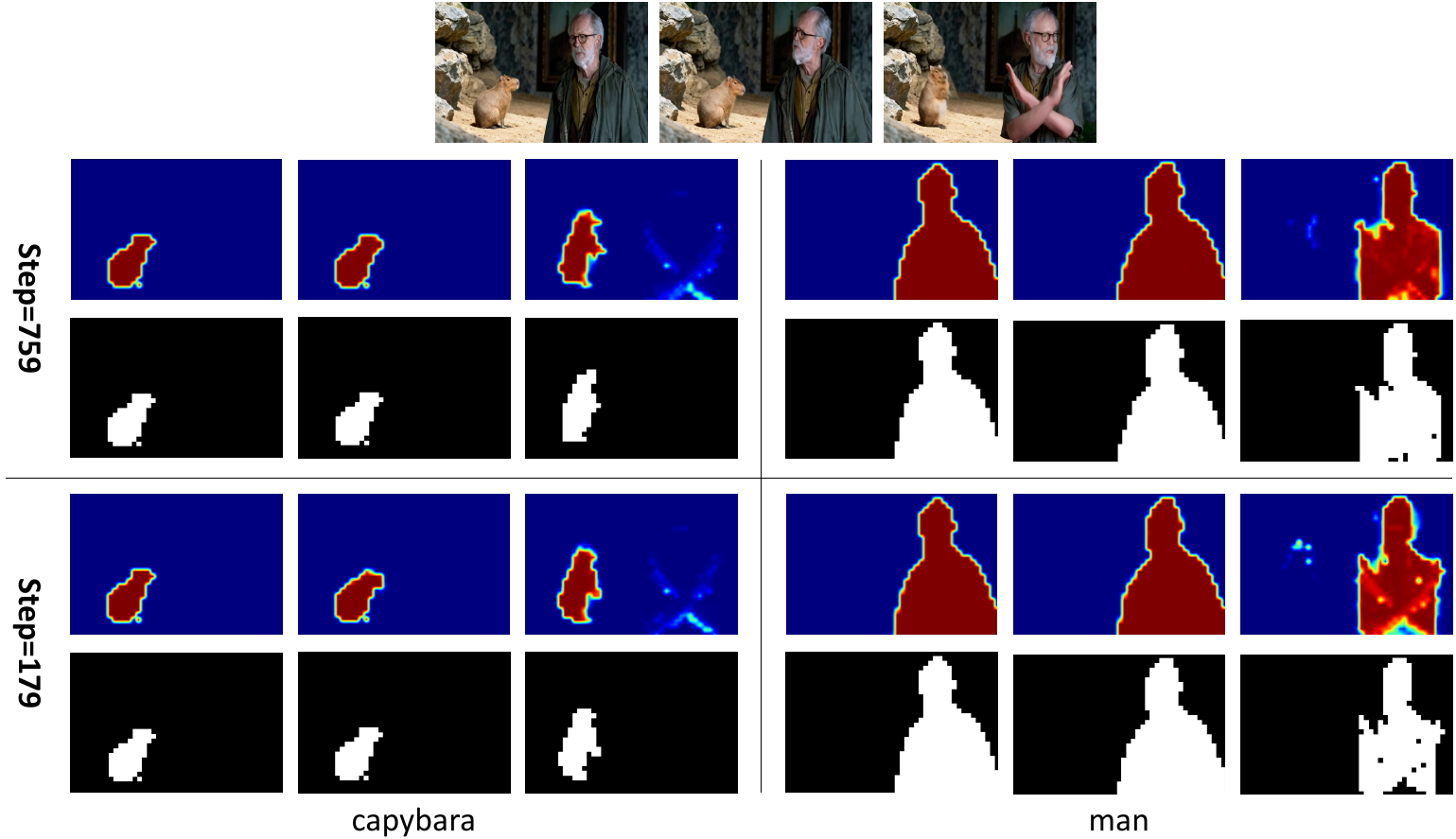}
    \caption{\textbf{Illustration of the propagate feature and mask changes during denoising steps.} We found that the transfer of motion can be completed in the early stage of denoising steps.}
    \label{fig:dynamic_extractioin}
    \vspace{-0.3cm}
\end{figure}
\myparagraph{Inference stage.}
At inference time, we should distinguish different objects in $I_\mathrm{tar}$. 
A straightforward solution is to reuse the simple attention-based mask extraction from the training stage. 
However, it fails to distinguish different objects,
as shown in \cref{fig:simple_extraction}, replecting that text-driven localization no longer provides reliable object when multiple objects appear in a single image.

To this end, we introduce a \textit{Regressive Mask Propagation Mechanism} (RMPM), which progressively propagates masks from the first frame to all subsequent frames, inspired by \cite{univst}.
We begin by using an off-the-shelf segmentation model \cite{groundedsam} to extract object-specific masks $\{\widehat{\mathcal{M}}^k_1\}_{k=1}^K$ for the first frame $I_\mathrm{tar}$.
We also extract latent features for all frames $\mathcal{F}=\{\mathcal{F}_l\}_{l=1}^L$, where $L$ is the number of frames.
For each object $o^k_\mathrm{tar}$, RMPM recursively propagates its mask from frame $1$ to frame $L$.
As the $l$-th frame, we maintain a small set of anchor features $\mathcal{F}_\mathrm{anc}$ and corresponding anchor masks $\widehat{\mathcal{M}}^k_\mathrm{anc}$.
These anchors include the first-frame and a subset of nearby frames within a local temporal window. 
We first compute the correlation matrix $\mathcal{C}^k_l$:
\begin{equation}
\mathcal{C}^k_l = \mathrm{Norm}(\mathcal{F}_l) \cdot \mathrm{Norm}(\mathcal{F}_\mathrm{anc} \odot \widehat{\mathcal{M}}^k_{\mathrm{anc}})^\top,
\label{eq:score_matrix}
\end{equation}
where $\mathcal{F}_\mathrm{anc}$ and $\widehat{\mathcal{M}}^k_\mathrm{anc}$ are constructed by concatenating the first-frame feature/mask $\mathcal{F}_1$/$\widehat{\mathcal{M}}^k_1$ with anchor features/masks.
$\mathrm{Norm}$ is the L2 normalization operation.

The propagated feature $\mathcal{S}^k_l$ and mask $\widehat{\mathcal{M}}^k_l$ for the $l$-th frame of the $k$-th object are then computed as:
\begin{equation}
\begin{split}
\mathcal{S}^k_l &= \mathcal{M}^k_\mathrm{anc} \cdot {\mathcal{C}^k_l}^\top, \\
\widehat{\mathcal{M}}^k_l &=
\begin{cases}
1,& \mathrm{if}~~ \mathcal{S}^k_l > \mathrm{mean}(\mathcal{S}^k_l)  \\
0,& \text{otherwise}
\end{cases}.
\end{split}
\label{eq:propogate_mask}
\end{equation}

After $\widehat{\mathcal{M}}^k_l$, we update the anchor set by appending the current frame’s feature and mask, removing the oldest ones when exceeding a predefined window size.
RMPM ensures that the mask for each object evolves consistently over time, enabling reliable separation of objects throughout the video.
The details of RMPM are presented in Algorithm \ref{alg:inf_mask}.

\begin{algorithm}[t]
	\begin{footnotesize}
		\SetAlgoLined
        \SetKwInOut{Input}{Input}
        \SetKwInOut{Output}{Output}
        \SetKwInput{Initialization}{Initialization}
        \SetKwInput{Training}{Training Process}
        \Input{First frame mask for $K$ objects: $\{\widehat{\mathcal{M}}^k_1\}_{k=1}^K$ ; Features for all frames $\mathcal{F}=\{\mathcal{F}_l\}^{L}_{l=1}$; Local temporal window size $W$; Object number $K$; Frame number $L$.
		}
		\Output{Object-specific masks $\{\widehat{\mathcal{M}}^k_l\}_{k=1, \ldots, K,l=1, \ldots, L}$.}
		\Initialization{Anchor features $\mathcal{F}_{\mathrm{anc}}$; Anchor masks for the $k$-th object $\widehat{\mathcal{M}}^k_{\mathrm{anc}}$.} 
		\For{$k=1$ \KwTo $K$}
		{
            $\mathcal{F}_{\mathrm{anc}}$ $\gets [~~]$; $\widehat{\mathcal{M}}^k_{\mathrm{anc}}$ $\gets [~~]$ \\
            \For{$l=1$ \KwTo $L$}
            {
                \If{$l=1$}{
                    $\widehat{\mathcal{M}}^k_l \gets \widehat{\mathcal{M}}^k_1$
                }
                \Else{
                    \If{$|\mathcal{F}_{\mathrm{anc}}| \geq W$}{
                        $\mathcal{F}_{\mathrm{anc}}.\text{pop}(1)$; $\widehat{\mathcal{M}}^k_{\mathrm{anc}}.\text{pop}(1)$
                    }
                    \Else{
                        Compute correlation matrix $\mathcal{C}^k_l$ via Eq.~\eqref{eq:score_matrix}; \\
                        Compute propagated mask $\widehat{\mathcal{M}}^k_l$ via Eq.~\eqref{eq:propogate_mask};
                    }
                }
                {$\mathcal{F}_{\mathrm{anc}}.\text{append}(\mathcal{F}_l)$; $\widehat{\mathcal{M}}^k_{\mathrm{anc}}.\text{append}(\widehat{\mathcal{M}}^k_l)$)}
            }
        }
		\caption{RMPM}\label{alg:inf_mask}
	\end{footnotesize}
\end{algorithm}
\myparagraph{Dynamic RMPM.} As pointed out in \cite{denoise1, denoise2}, early denoising steps in diffusion models primarily recover coarse structural information.
Similarly, in our RMPM, we observe that applying mask propagation during the early denoising steps is sufficient for accurate mask estimation, as illustrated in \cref{fig:dynamic_extractioin}.
This motivates us to propose a dynamic RMPM, a more efficient variant of RMPM.
Specifically, at each diffusion step $t$, we compute the difference between the current mask and the mask from the previous step $t-1$. If this difference falls below a pre-defined threshold $\alpha$, we terminate further mask updates and reuse the most recent stable mask for all subsequent steps. This significantly improves inference efficiency without compromising mask quality.

\section{Experiment}
\label{sec:experiment}
\begin{table*}[t]
    \centering
    \caption{\textbf{Quantitative comparison.} We compared the effectiveness of different methods with our method. There are five metrics for Automatic Evaluations and four for Human Evaluations, among which Human Evaluations score are in percentage.}
    \label{tab:comparison}
    \setlength{\tabcolsep}{9pt}
    \renewcommand{\arraystretch}{1.1}
    \scalebox{0.9}
    {
    \begin{tabular*}{\textwidth}
    {l|ccccc|cccc}
        \hline
        \multicolumn{1}{l|}{\textbf{\multirow{2}{*}{Method}}} & \multicolumn{5}{c|}{Automatic Evaluations} & \multicolumn{4}{c}{Human Evaluations (\%)}\\ 
        \cmidrule(r{8pt}){2-10}
         & \textbf{AC $\uparrow$} & \textbf{TC $\uparrow$} & \textbf{TS $\uparrow$} & \textbf{TF $\uparrow$} & \textbf{FF $\uparrow$} & \textbf{AC $\uparrow$} & \textbf{TC $\uparrow$} & \textbf{TS $\uparrow$} & \textbf{MF $\uparrow$} \\ 
        \hline
        AnyV2V \cite{anyv2v} & \textcolor{lightgray}{0.749} & \textcolor{lightgray}{0.890} & \textcolor{lightgray}{0.275} & 0.422 & 0.562 & 0.000 & 0.000 & 0.000 & 0.000\\ 
        Flexiact \cite{flexiact} & \textcolor{lightgray}{0.939} & \textcolor{lightgray}{0.954} & \textcolor{lightgray}{0.281} & 0.279 & 0.609 & 15.750 & 11.550 & 6.475 & 6.500\\
        I2VEdit \cite{i2vedit} & \textcolor{lightgray}{0.915} & \textcolor{lightgray}{0.932} & \textcolor{lightgray}{0.267} & 0.395 & 0.521 & 2.000 & 0.575 & 0.000 & 0.000\\ 
        Go-with-the-Flow \cite{go-with-the-flow} & \textcolor{lightgray}{0.774} & \textcolor{lightgray}{0.881} & \textcolor{lightgray}{0.277} & 0.488 & 0.648 & 0.000 & 0.000 & 0.500 & 0.525\\ 
        CogVideoX-5B-I2V \cite{cogvideox} & \textcolor{lightgray}{0.948} & \textcolor{lightgray}{0.956} & \textcolor{lightgray}{0.293} & 0.292 & 0.607 & 5.500 & 4.000 & 3.475 & 3.500\\ 
        \hline
        \textbf{Our \Our}& \textcolor{lightgray}{\textbf{0.904}}& \textcolor{lightgray}{\textbf{0.932}}& \textcolor{lightgray}{\textbf{0.286}}& \textbf{0.577} & \textbf{0.723} & \textbf{76.750}& \textbf{83.875} & \textbf{89.550} & \textbf{89.475}\\ 
        \hline
    \end{tabular*}
    }
\end{table*}
\begin{figure*}
    \centering
    \includegraphics[width=\textwidth]{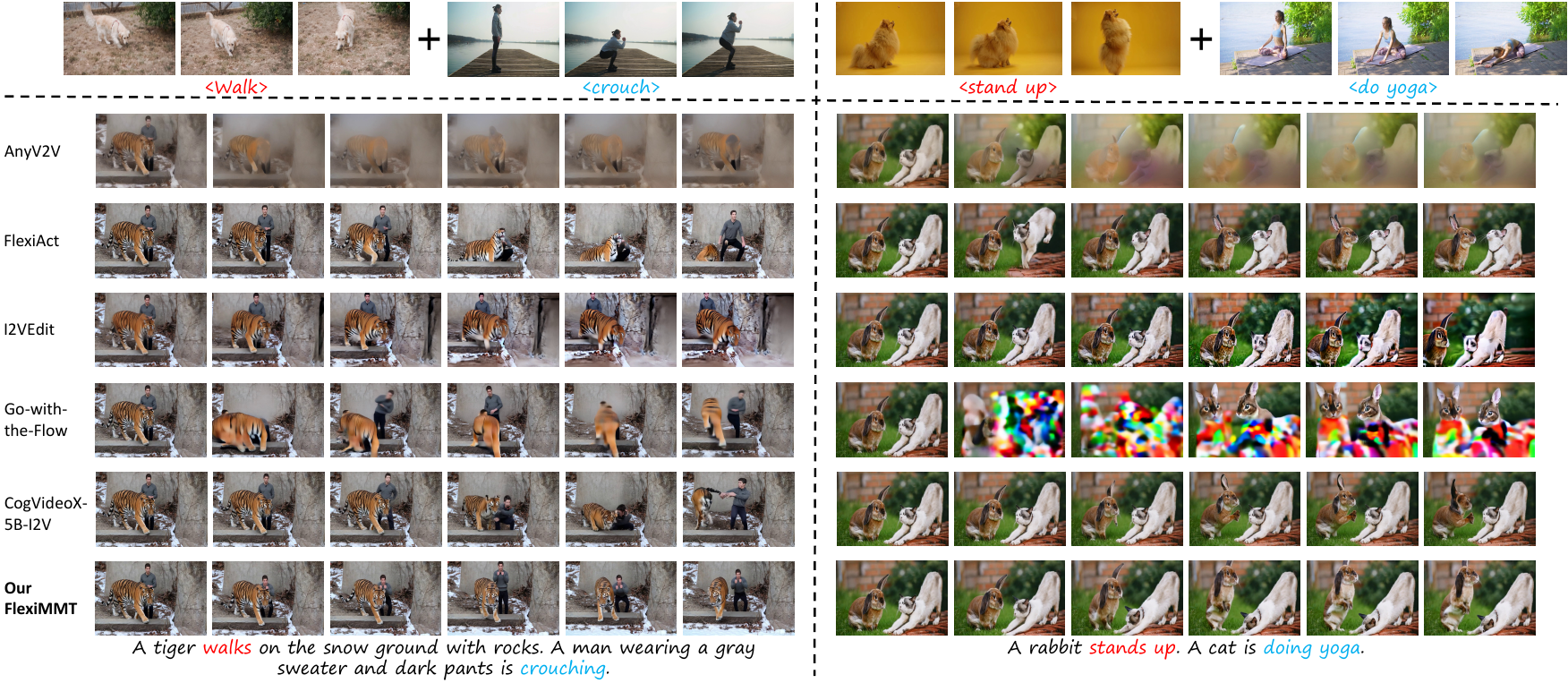}
    \caption{\textbf{Qualitative comparison.} We compared the effects of different methods on transferring multi-object motions. The first row is the referenced motion, and the next six rows are the results of different methods.}
    \label{fig:qualitative}
    \vspace{-0.5cm}
\end{figure*}

\subsection{Experimental Settings}

\myparagraph{Datasets.}
To ensure diversity, we collect and generate paired videos and images from multiple sources, including FlexiAct \cite{flexiact}, Pexels\footnote{\url{https://www.pexels.com/}}, the DAVIS 2017 dataset \cite{davis2017}, and Seedream \cite{seedream}.
In total, we construct a dataset containing 200 video–image pairs covering 20 distinct types of motion. Each image contains single or multi objects, enabling evaluation under multi-object multi-motion transfer scenarios.

\myparagraph{Baselines.}
We compare our approach against several representative baselines, including FlexiAct \cite{flexiact}, I2VEdit \cite{i2vedit}, AnyV2V \cite{anyv2v}, Go-with-the-Flow \cite{go-with-the-flow}, and CogVideoX-5B-I2V \cite{cogvideox}, which together cover both first-frame-guided and multimodal-guided paradigms.
For a fair comparison, we adapt each baseline to enable multi-object, multi-motion transfer. Details of these modifications are provided in the Supplementary Material.

\myparagraph{Evaluation metrics.}
For automatic evaluations, following \cite{spacetimediffusion, vmc, flexiact}, we evaluate generated videos with four metrics:
Appearance Consistency (AC), Temporal Consistency (TC), Text Similarity (TS), and Trajectory Fidelity (TF)\footnote{Referred to as Motion Fidelity in FlexiAct \cite{flexiact}}, measuring appearance preservation, temporal stability, semantic alignment, and motion coherence, respectively.

To better quantify motion alignment, we further introduce a new metric, \textit{Flow Fidelity} (FF).
We use the RAFT \cite{raft} optical flow estimator to compute the similarity between the flow fields of corresponding objects in the generated and reference videos.
FF evaluates both the similarity of motion-magnitude distributions and the histogram similarity of motion directions within the object masks.
Compared with TF, FF can measure the similarity of motion from a more holistic dimension.

For human evaluations, we adopt four criterias, \ie, Appearance Consistency (AC), Temporal Consistency (TC), Text Similarity (TS), Motion Fidelity (MF), to capture appearance preservation, temporal stability, semantic alignment, and motion coherence. We recruited 20 raters, each randomly assigned 200 video sequences. For each criterion, raters were asked to select the best-performing video. Additional details are provided in the Supplementary Material.

\myparagraph{Implementation details.}
We adopt CogVideoX-5B-I2V~\cite{cogvideox} as our base I2V diffusion model.
All videos and images are resized to $720 \times 480$, and each video contains 49 frames.
Motion tokens is trained for 2,000 steps using AdamW optimizer with a learning rate of 3e-3 and batch size of 1.
We use \cite{groundedsam} for first-frame mask extraction.
Additionally, we set the local temporal window size to $W = 2$ in RMPM and the threshold to $\alpha = 5\%$ in dynamic RMPM.
All experiments are conducted on six NVIDIA A800 GPUs.

\subsection{Main Results}
\myparagraph{Quantitative Evaluation.}
As shown in ~\cref{tab:comparison}, our \Our performs the best in both Trajectory Fidelity (TF) and Flow Fidelity (FF), demonstrating its superior ability to accurately transfer motion.
Compared with \cite{go-with-the-flow}, which achieves the best results on TF and FF among all baselines, our method further obtains higher Appearance Consistency (AC), Temporal Consistency (TC), and Text Similarity (TS). This indicates that our \Our not only preserves motion most faithfully but also maintains superior overall generation quality.

It is noting that our \Our achieves comparable or slightly lower performance on AC, TC and TS compared with previous methods~\cite{flexiact,i2vedit,cogvideox}. The underlying reason is that AC and TC measure inter-frame similarity using CLIP~\cite{clip}. These metrics favor videos with minimal changes and therefore become uninformative when a model fails to transfer motion.
That is to say, high AC/TC may simply indicate static or weakly animated outputs, reflecting by TF and FF, and qualitative and human evaluation. 
Similarly, TS cannot reliably reflect motion accuracy because text descriptions inadequately express fine-grained motion details.
Hence, these quality metrics are meaningful only when motion fidelity is sufficiently high.
Overall, \Our shows the strongest motion transfer capability while maintaining high and comparable generation quality.

\myparagraph{Human evaluation.} The right side of ~\cref{tab:comparison} shows that our method achieves the best performance across all human-evaluated metrics.

\myparagraph{Qualitative results.}
Results in ~\cref{fig:qualitative} shows our qualitative results. It can be seen that none of the existing methods can transfer the motion of multiple objects. In contrast, our method faithfully transfers: ``walk'' motion to tiger and ``crouch'' motion to man, ``stand up'' motion to rabbit and ``do yoga'' motion to cat.

\begin{table}[t]
    \centering
    \caption{\textbf{Impact of M2X and T2X masks in MDMA.} ``w/o M2X'' denotes removing all Motion-to-[X] masks and ``w/o T2X'' denotes removing all Text-to-[X] masks during inference.}
    \label{tab:ablation}
    \setlength{\tabcolsep}{6pt}
    \renewcommand{\arraystretch}{1.1}
    \scalebox{0.9}
    {
    \begin{tabular}{l|ccccc}
        \hline
         \textbf{Method} & \textbf{AC $\uparrow$} & \textbf{TC $\uparrow$} & \textbf{TS $\uparrow$} & \textbf{TF $\uparrow$} & \textbf{FF $\uparrow$}\\ 
        \hline
        w/o M2X & \textcolor{lightgray}{0.910} & \textcolor{lightgray}{0.940} & \textcolor{lightgray}{0.284} & 0.381 & 0.618\\ 
        w/o T2X & \textcolor{lightgray}{0.926} & \textcolor{lightgray}{0.946} & \textcolor{lightgray}{0.284} & 0.461 & 0.665\\
        \hline
        \textbf{Our \Our}& \textcolor{lightgray}{0.904}& \textcolor{lightgray}{0.932}& \textcolor{lightgray}{0.286}& \textbf{0.577} & \textbf{0.723}\\ 
        \hline
    \end{tabular}
    }
\end{table}

\begin{figure}[t]
    \centering
    \includegraphics[width=\linewidth]{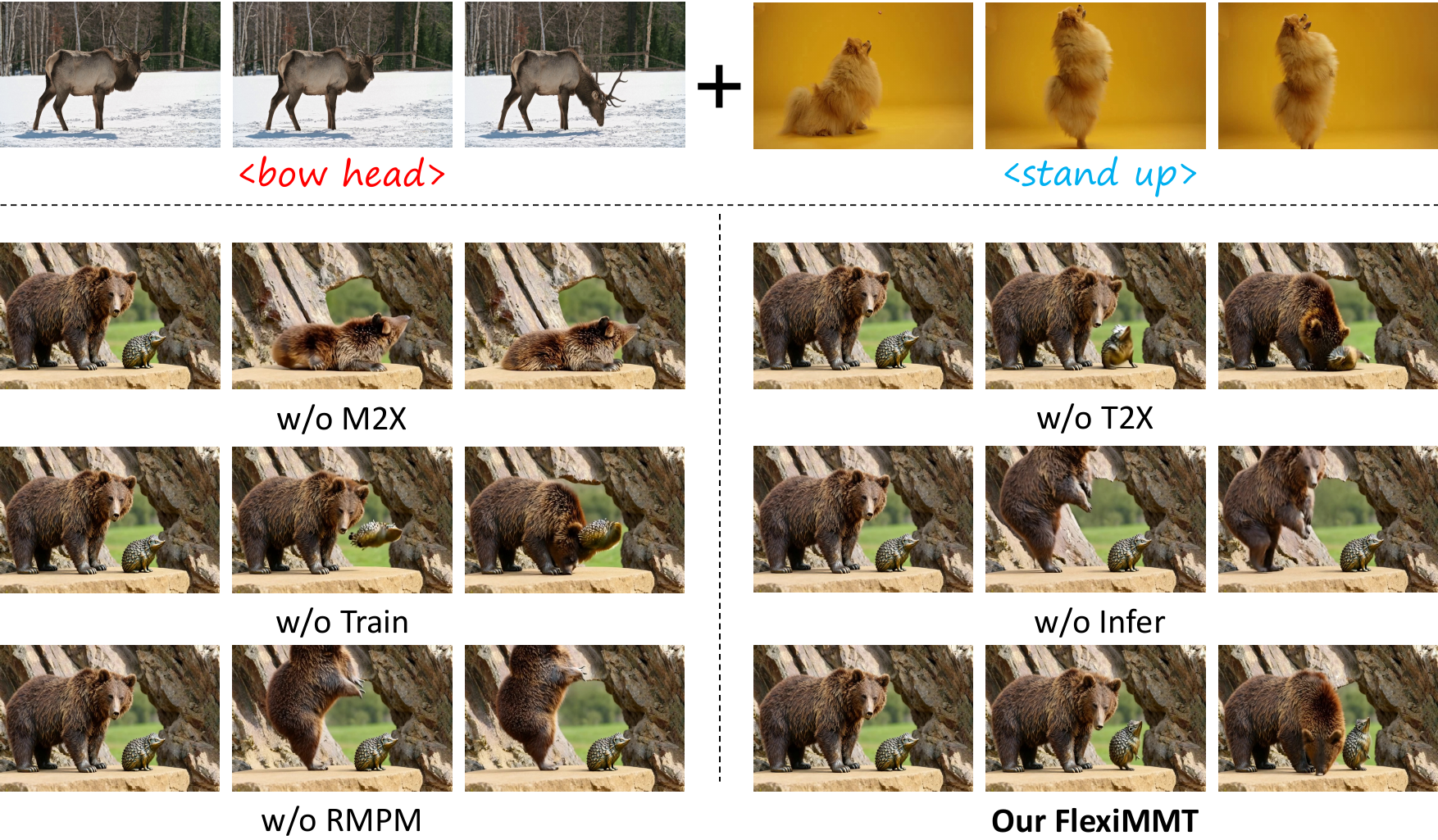}
    \caption{\textbf{Qualitative comparison of each component in \Our.} The caption for the generated videos is: ``A beer bows head. A hedgehog stands up.''}
    \label{fig:ablation}
    \vspace{-0.3cm}
\end{figure}

\subsection{Ablation Studies}

\myparagraph{Impact of M2X and T2X masks in MDMA.}
As shown in the upper block of ~\cref{tab:ablation}, removing either the M2X or T2X mask leads to a significant drop in TF and FF, indicating that both components are essential for effective motion disentanglement across objects.
~\cref{fig:ablation} further illustrates that, without M2X and T2X, motion cannot be correctly transferred to multiple objects.
More ablations for each part of M2X and T2X masks are in Supplementary Material.

\myparagraph{Impact of DMEM.} DMEM includes both training-stage and inference-stage mask extraction strategies.
As shown in the first two rows of ~\cref{tab:ablation_DMEM}, both stages are crucial for effective motion transfer.
The qualitative results in ~\cref{fig:ablation} further illustrate that: i) without the training-stage mask, the object fails to accurately learn the reference motion;
ii) without the inference-stage mask, multiple motions become entangled.

\begin{table}[t]
    \centering
    \caption{\textbf{Impact of DMEM and RMPM.} ``w/o Train'' and  ``w/o Infer'' indicate disabling mask extraction during training and inference, respectively. ``w/o RMPM'' denotes removing the regressive mask extraction mechanism (RMPM) at inference.
    }
    \label{tab:ablation_DMEM}
    \setlength{\tabcolsep}{6pt} 
    \renewcommand{\arraystretch}{1.1}
    \scalebox{0.9}
    {
    \begin{tabular}{l|ccccc}
        \hline
         \textbf{Method} & \textbf{AC $\uparrow$} & \textbf{TC $\uparrow$} & \textbf{TS $\uparrow$} & \textbf{TF $\uparrow$} & \textbf{FF $\uparrow$}\\ 
        \hline
        w/o Train & \textcolor{lightgray}{0.925} & \textcolor{lightgray}{0.945} & \textcolor{lightgray}{0.285} & 0.440 & 0.656\\ 
        w/o Infer & \textcolor{lightgray}{0.884} & \textcolor{lightgray}{0.924} & \textcolor{lightgray}{0.285} & 0.373 & 0.602\\
        w/o RMPM & \textcolor{lightgray}{0.895} & \textcolor{lightgray}{0.924} & \textcolor{lightgray}{0.285} & 0.377 & 0.607\\
        \hline
        \textbf{Our \Our}& \textcolor{lightgray}{0.904}& \textcolor{lightgray}{0.932}& \textcolor{lightgray}{0.286}& \textbf{0.577} & \textbf{0.723}\\ 
        \hline
    \end{tabular}
    }
\end{table}

\begin{figure}[t!]
    \centering
    \includegraphics[width=0.9\linewidth]{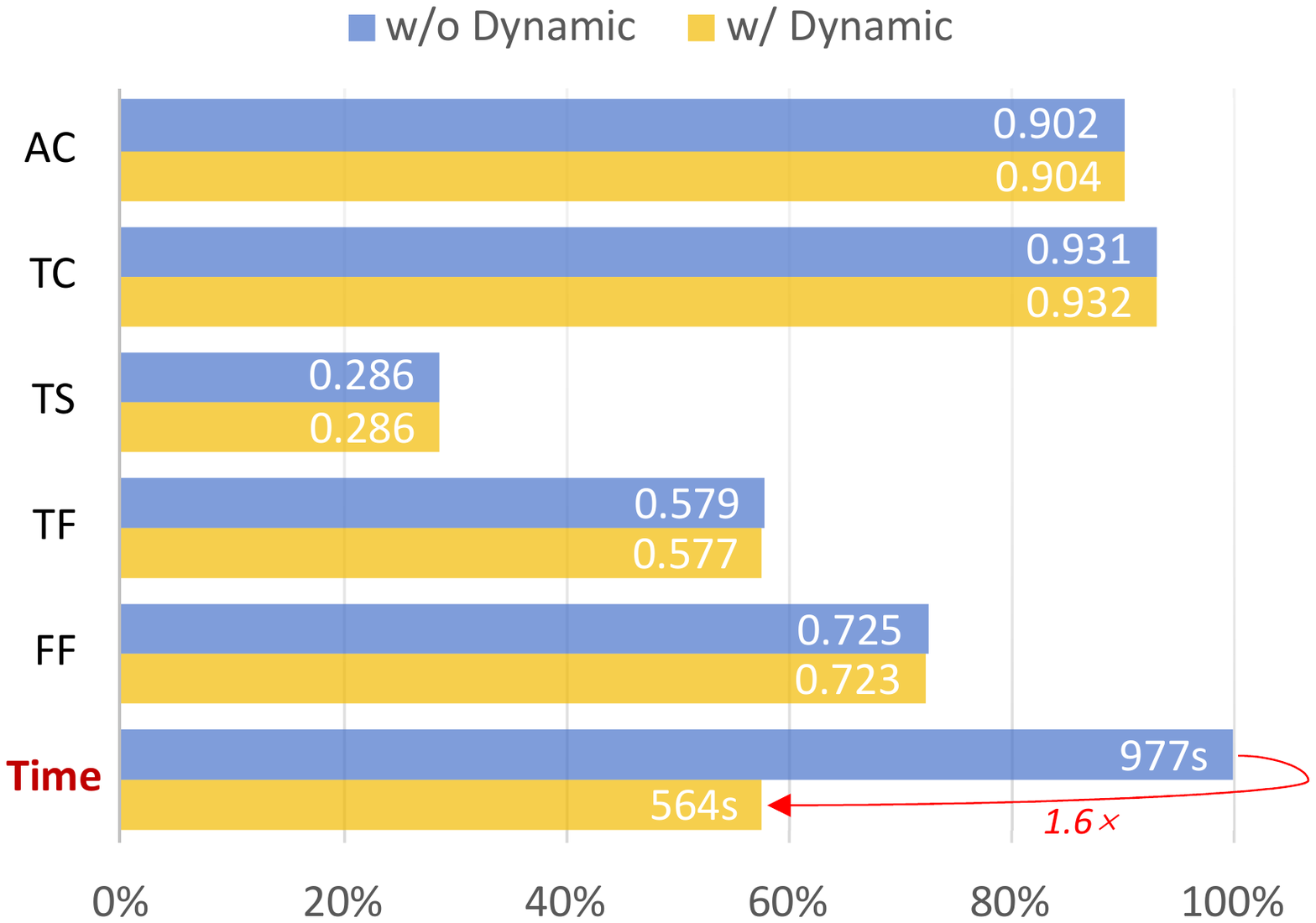}
    \caption{\textbf{Comparison of RMPM (w/o Dynamic) and Dynamic RMPM (w/ Dynamic).} Dynamic RMPM substantially accelerates inference without compromising performance.}
    \label{fig:DME}
    \vspace{-0.4cm}
\end{figure}

\myparagraph{Impact of RMPM.} RMPM is the key inference-stage component in DMEM for achieving object-specific masks. Without RMPM, the model reduced to using only the simple training-stage mask extraction during inference.
As shown in the third row of  ~\cref{tab:ablation_DMEM} and in the visualizations of ~\cref{fig:ablation}, removing RMPM leads to entangled motions across objects, causing the model to fail in accurately following the prescribed motion patterns.

\myparagraph{RMPM \vs Dynamic RMPM.}
We tested the impact of Dynamic RMPM. As shown in \cref{fig:DME}, Dynamic RMPM significantly reduces the inference time.
\section{Conclusion}
\label{sec:conclusion}
In this paper, we propose \Our, the first implicit I2V motion transfer framework capable of handling multi-object, multi-motion scenarios. 
We introduce a motion decoupled mask attention mechanism and a differentiated mask extraction mechanism to effectively resolve cross-object motion entanglement issue, ensuring precise, independent motion assignment to each object. 
The progressive and dynamic mask extraction strategies further ensure stable multi-object control and efficient inference. Extensive experiments demonstrate that our \Our offers flexible, compositional, and high-fidelity motion transfer across diverse objects.

{
    \small
    \bibliographystyle{ieeenat_fullname}
    \bibliography{main}
}

\clearpage
\setcounter{page}{1}
\maketitlesupplementary

\setcounter{section}{0}
\renewcommand{\thesection}{\Alph{section}}
\section{More Details}

\subsection{Full Text Prompts in ~\cref{fig:big_image}}
For the left-top image containing a single object with the motion \textcolor{red}{$<$do yoga$>$}, the full text prompt is: ``A man is doing yoga.''

For the right-top image containing 2 objects with motions \textcolor{darkblue}{$<$raise hands$>$} and \textcolor{darkgreen}{$<$exercise$>$}, the full text prompt is: ``A female cartoon character in armor raised his hands above his head while placing his right foot against his left knee. A shirtless man with a muscular build and a beard is performing a fitness exercise.''

For the left-bottom image containing 3 objects with motions \textcolor{darkgreen}{$<$exercise$>$}, \textcolor{red}{$<$do yoga$>$}, and \textcolor{darkblue}{$<$raise hands$>$}, the full text prompt is: ``The left man is performing a fitness exercise. The middle man is doing yoga. The right man raised his hands above his head while placing his right foot against his left knee.''

For the right-bottom image containing 3 objects with motions \textcolor{red}{$<$do yoga$>$}, \textcolor{darkgreen}{$<$exercise$>$}, and \textcolor{darkblue}{$<$raise hands$>$}, the full text prompt is: ``The left man is doing yoga. The middle man is performing a fitness exercise. The right man raised his hands above his head while placing his right foot against his left knee.''
Note that the order of motions is different from the above one, showing that our \Our is able to support flexible, compositional, and arbitrary motion–object rearrangements.

\subsection{Details of Baseline Modifications}
We evaluate five representative image-to-video (I2V) baselines, including FlexiAct \cite{flexiact}, I2VEdit \cite{i2vedit}, AnyV2V \cite{anyv2v}, Go-with-the-Flow \cite{go-with-the-flow}, and CogVideoX-5B-I2V \cite{cogvideox}.
To ensure a fair comparison and enable multi-object, multi-motion transfer across all methods without altering their core contributions, we introduce minimal adjustments to their implementations. The details are as follows:

\myparagraph{AnyV2V~\cite{anyv2v}.} We aggregate information from multiple source videos by averaging their inverted noise and intermediate feature representations to support multi-object multi-motion transfer.

\myparagraph{FlexiAct~\cite{flexiact}.} FlexiAct trains an individual Freq-aware Embedding (Motion Tokens in our paper) for each motion from a single object. To enable multi-object, multi-motion transfer, we simply concatenate multiple motion tokens extracted from different reference videos.

\myparagraph{I2VEdit~\cite{i2vedit}.} I2VEdit trains a distinct LoRA module to represent each specific motion. Accordingly, during inference, we enable multi-object, multi-motion transfer by averaging and average the LoRA weights corresponding to different motions.

\myparagraph{Go-with-the-Flow~\cite{go-with-the-flow}.} For Go-with-the-Flow, we use its first-frame editing (I2V) functionality. To achieve multi-object multi-motion transfer, we first align object features from the source video’s initial frame to their corresponding locations in the target image. We then apply geometric transformations derived from the source masks to the target masks to localize each object, and use the extracted flow fields to guide the motion.

\myparagraph{CogVideoX-5B-I2V~\cite{cogvideox}.} CogVideoX-5B-I2V is designed solely for image-to-video generation and does not inherently support motion transfer. Therefore, for this baseline, we simply use the first frame and its corresponding caption to generate the video.
In contrast, our \Our is built upon CogVideoX-5B-I2V and equips it with the capability for multi-object, multi-motion transfer.

\begin{figure*}[htbp]
    \centering
    \includegraphics[width=\linewidth]{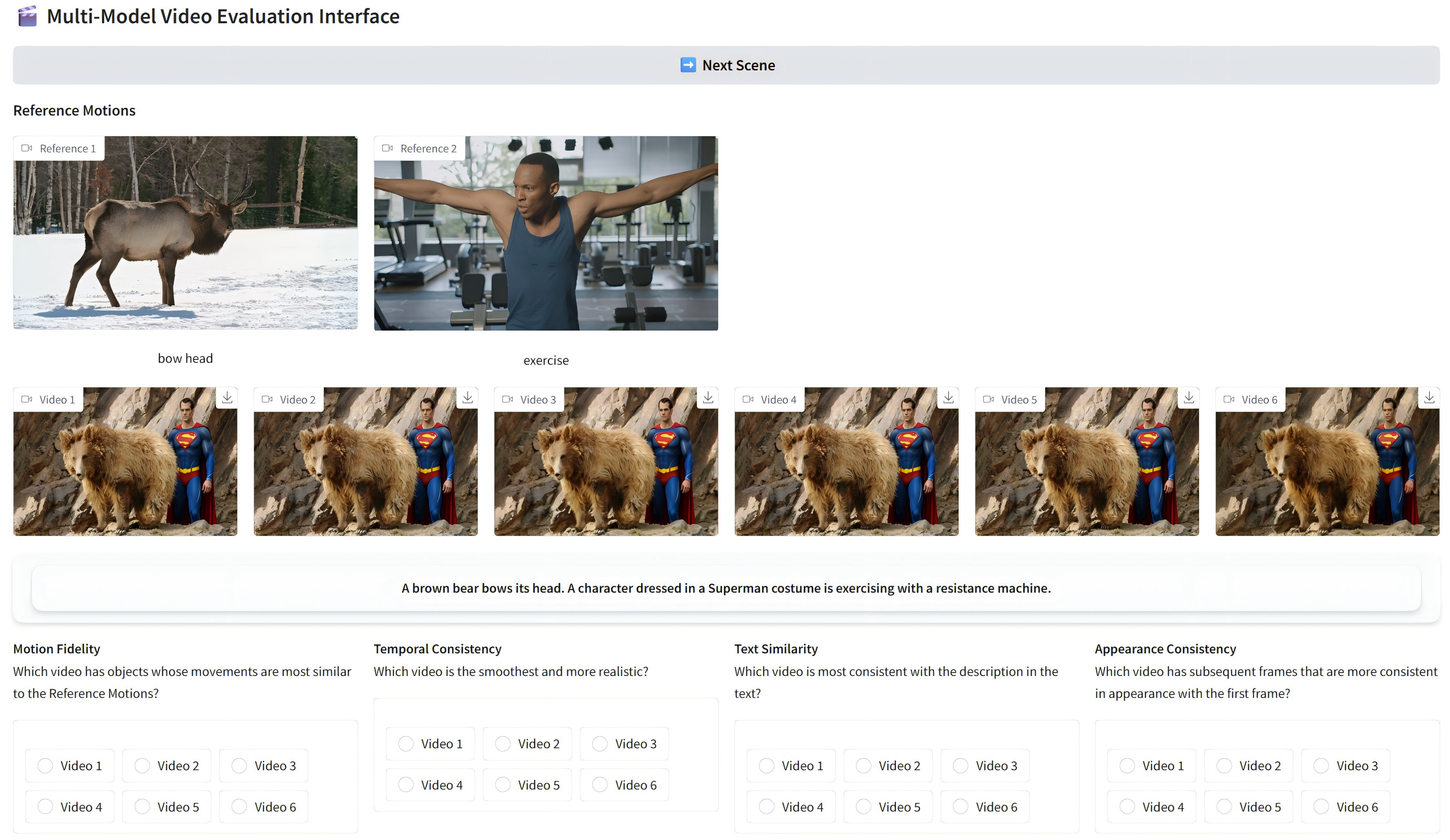}
    \caption{Visulization of human evaluation interface}
    \label{fig:interface}
\end{figure*}
\subsection{Details of Human Evaluation}
As shown in \cref{fig:interface}, we use the Gradio\footnote{Abid, Abubakar, Ali Abdalla, Ali Abid, Dawood Khan, Abdulrahman Alfozan, and James Zou. ``Gradio: Hassle-free sharing and testing of ML models in the wild.'' arXiv preprint arXiv:1906.02569 (2019).} toolbox to build a web interface that allows raters to perform annotation. In our human evaluation, we recruited 20 raters, each of whom was assigned 200 video sequences presented in random order. The interface first displays the reference motions, corresponding to the reference videos, followed by 6 video pairs generated by the five baseline models and our \Our using the same initial frame, caption, and reference motions. For fairness, we shuffled the order of the videos, and each is anonymized as the $N$-th candidate  video. At the bottom of the interface, raters select the best video among the six candidates for each of the four evaluation indicators, including Appearance Consistency (AC), Temporal Consistency (TC), Text Similarity (TS), and Motion Fidelity (MF), through a single-choice scoring module.
The definitions of these four indicators are provided in \cref{fig:human_prefer_ui}.

\begin{figure*}[t]
\begin{tcolorbox}[colback=yellow!8, colframe=black, boxrule=0.8pt, arc=8pt, left=2pt, right=2pt, top=4pt, bottom=4pt, width=\textwidth]
\begin{itemize}[left=1em]
    \item \textbf{Motion Fidelity (MF)}: Which video has objects whose movements are most similar to the Reference Motions?
    \item \textbf{Temporal Consistency (TC)}: Which video is the smoothest and more realistic?
    \item \textbf{Text Similarity (TS)}: Which video is most consistent with the description in the text?
    \item \textbf{Appearance Consistency (AC)}: Which video has subsequent frames that are more consistent in appearance with the first frame?
\end{itemize}
\end{tcolorbox}
\caption{Detail of four indicators.}
\label{fig:human_prefer_ui}
\end{figure*}

\section{Additional Experiments}
\begin{table}[htbp]
    \centering
    \caption{Impact of each mask part in MDMA.}
    \setlength{\tabcolsep}{8pt}
    \renewcommand{\arraystretch}{1.1}
    \scalebox{0.9}
    {
    \begin{tabular}{l|ccccc}
        \hline
         \textbf{Method} & \textbf{AC $\uparrow$} & \textbf{TC $\uparrow$} & \textbf{TS $\uparrow$} & \textbf{TF $\uparrow$} & \textbf{FF $\uparrow$}\\ 
        \hline
        w/o M2V & \textcolor{lightgray}{0.907} & \textcolor{lightgray}{0.938} & \textcolor{lightgray}{0.285} & 0.383 & 0.616\\ 
        w/o M2M & \textcolor{lightgray}{0.902} & \textcolor{lightgray}{0.931} & \textcolor{lightgray}{0.286} & 0.576 & 0.723\\  
        w/o T2V & \textcolor{lightgray}{0.900} & \textcolor{lightgray}{0.931} & \textcolor{lightgray}{0.287} & 0.572 & 0.720\\ 
        w/o T2T & \textcolor{lightgray}{0.903} & \textcolor{lightgray}{0.931} & \textcolor{lightgray}{0.286} & 0.573 & 0.722\\ 
        w/o T2M & \textcolor{lightgray}{0.928} & \textcolor{lightgray}{0.948} & \textcolor{lightgray}{0.284} & 0.476 & 0.663\\ 
        \hline
        \textbf{Ours}& {0.904}& {0.932}& {0.286}& \textbf{0.577} & \textbf{0.723}\\ 
        \hline
    \end{tabular}
    }
    \label{tab:suppl_inf_mask}
\end{table}

\myparagraph{Impact of each mask part in MDMA.}
Our MDMA framework consists of two major components, \ie, Motion-to-[X] (M2X) mask and Text-to-[X] (T2X) mask.
The M2X masks ensure that motion-related tokens focus only on the specified object. This includes the Motion-to-Video and its symmetric Video-to-Motion masks (collectively referred to as M2V), as well as the Motion-to-Motion (M2M) mask.
The T2X masks guarantee that motion-related text tokens attend only to the video tokens corresponding to their associated object, thereby preventing cross-object motion interference. This category includes the Text-to-Video and Video-to-Text (T2V), Text-to-Text (T2T), and Text-to-Motion and Motion-to-Text (T2M) masks.

We conduct detailed ablation studies on each component of MDMA, as shown in \cref{tab:suppl_inf_mask}.
Using all mask components yields consistently strong performance across metrics, particularly in Trajectory Fidelity (TF) and Flow Fidelity (FF), demonstrating that every component contributes to accelerating multi-object, multi-motion transfer.
There is a slight decrease in Text Similarity (TS) compared with the w/o T2V setting (0.286 vs. 0.287). This is because non-motion tokens in the text description may provide additional cues that enhance text–video alignment, but at the same time introduce motion entanglement that degrades TF and FF.
The w/o M2V and w/o T2M settings yield higher Appearance Consistency (AC) and Temporal Consistency (TC); however, they severely impair TF and FF. This occurs because these settings mix motion tokens across different objects, and in some cases even produce static or low-motion videos—leading to artificially high AC and TC but extremely low TF and FF.
Given the objective of multi-object, multi-motion transfer, we adopt all mask components as our final \Our.

\begin{table}[htbp]
    \centering
    \caption{Anchor frames $W$ in RMPM. Time represents the time required for each step of denoising.}
    \setlength{\tabcolsep}{8pt}
    \renewcommand{\arraystretch}{1.1}
    \scalebox{0.9}
    {
    \begin{tabular}{l|cccccc}
        \hline
         $W$ & \textbf{AC $\uparrow$} & \textbf{TC $\uparrow$} & \textbf{TS $\uparrow$} & \textbf{TF $\uparrow$} & \textbf{FF $\uparrow$} & \textbf{Time}\\ 
        \hline
        1 & 0.901 & 0.931 & 0.288 & 0.575 & 0.718 & 17.39s\\ 
        3 & \textbf{0.904} & \textbf{0.932} & \textbf{0.286} & \textbf{0.577} & \textbf{0.723} & \textbf{17.81s}\\  
        5 & 0.903 & 0.932 & 0.287 & 0.582 & 0.724 & 18.60s\\ 
        \hline
    \end{tabular}
    }
    \label{tab:suppl_m}
\end{table}

\paragraph{Effect of $W$ in RMPM.}
In RMPM, we maintain a small set of anchor features and corresponding anchor masks to calculate the correlation matrix.
We use a local temporal window size $W$ to control the number of anchor frames included. In this subsection, we evaluate the impact of varying the number of anchor frames.
The results are shown in \cref{tab:suppl_m}.
When $W=1$, our method drops to only using the first frame as the anchor frame, \ie, the anchor set contains only the initial frame.
Increasing $W$ to $3$, \ie, using the two nearest frames in addition to the first frame, improves performance across all metrics, although training efficiency decreases slightly.
Further increasing $W$ yields marginal additional performance gains but leads to higher inference cost. 
Considering the trade-off between effectiveness and efficiency, we set $W$ to $2$ in all experiments by default.

\begin{table}[htbp]
    \centering
    \caption{Effect of $\alpha$ in Dynamic RMPM.}
    \setlength{\tabcolsep}{8pt}
    \renewcommand{\arraystretch}{1.1}
    \scalebox{0.9}
    {
    \begin{tabular}{l|cccccc}
        \hline
         $\alpha$ & \textbf{AC $\uparrow$} & \textbf{TC $\uparrow$} & \textbf{TS $\uparrow$} & \textbf{TF $\uparrow$} & \textbf{FF $\uparrow$} & \textbf{Time}\\ 
        \hline
        w/o & 0.902 & 0.931 & 0.286 & 0.579 & 0.725 & 977s\\ 
        5\% & \textbf{0.904} & \textbf{0.932} & \textbf{0.286} & \textbf{0.577} & \textbf{0.723} & \textbf{564s}\\  
        10\% & 0.902 & 0.930 & 0.286 & 0.577 & 0.722 & 530s\\ 
        \hline
    \end{tabular}
    }
    \label{tab:suppl_alpha}
\end{table}

\paragraph{Effect of $\alpha$ in Dynamic RMPM.}
In Dynamic RMPM, $\alpha$ is a pre-defined threshold used to determine mask stability.
If the difference between the current mask and the mask from the previous step falls below $\alpha$, further mask updates are terminated, and the most recent stable mask is reused for all subsequent steps.

In \cref{tab:suppl_alpha}, we evaluate the impact of different $\alpha$ values. The results show that $\alpha=5\%$ yields the best overall performance while maintaining balanced computational efficiency. Thus, we set $\alpha$ to $5$ in all experiments by default.

\begin{figure*}[t]
    \centering
    \includegraphics[width=\linewidth]{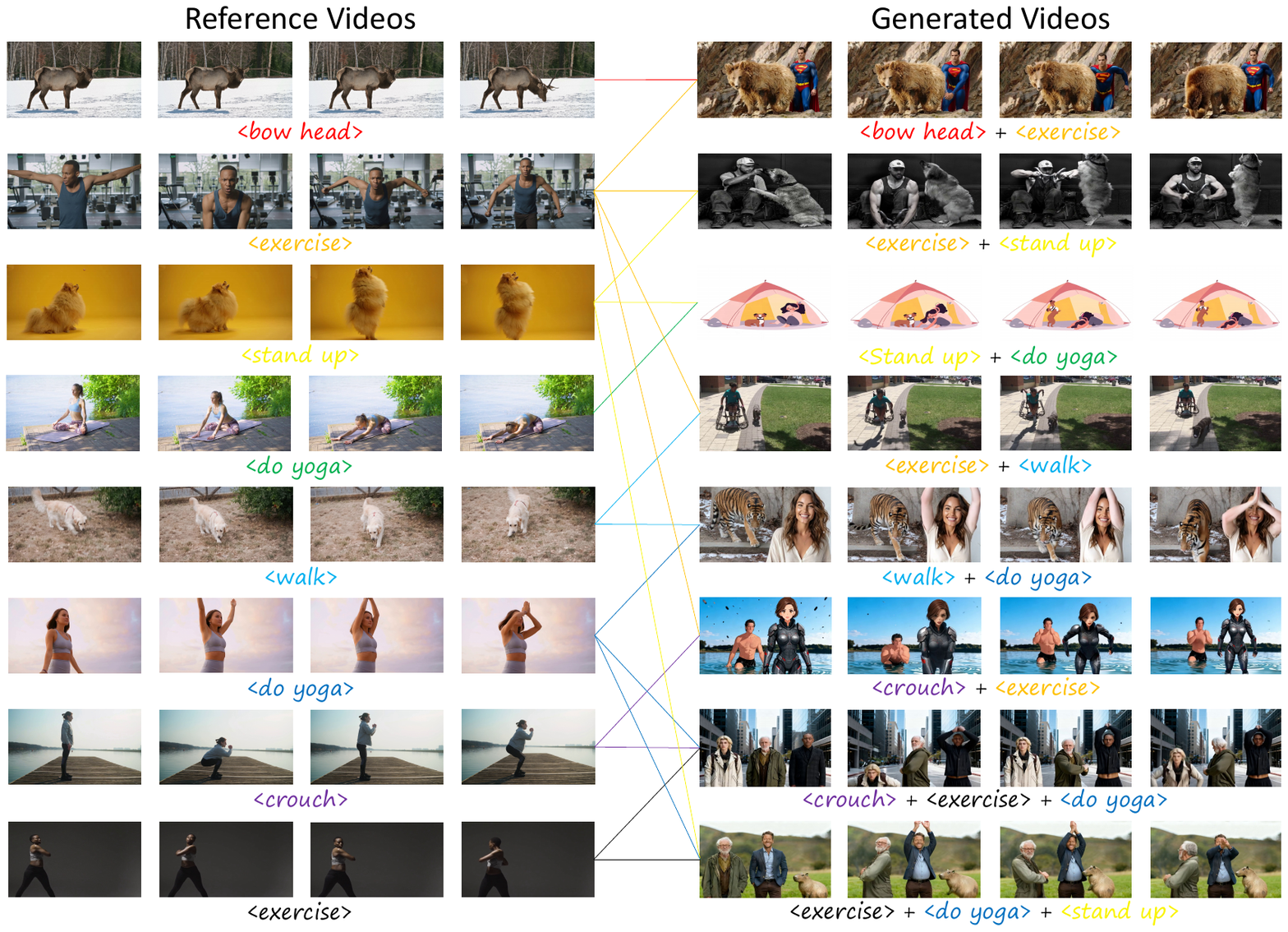}
    \caption{\textbf{More qualitative comparisons.} The left row shows reference videos. The right row shows generated videos.}
    \label{fig:additional_res}
\end{figure*}

\myparagraph{More results.}
\cref{fig:additional_res} presents additional qualitative results. The results
demonstrate that our method can support flexible, compositional, and arbitrary motion–object rearrangements, effectively handling arbitrary multi-object, multi-motion transfer scenarios.

\begin{figure}[t]
    \centering
    \includegraphics[width=\linewidth]{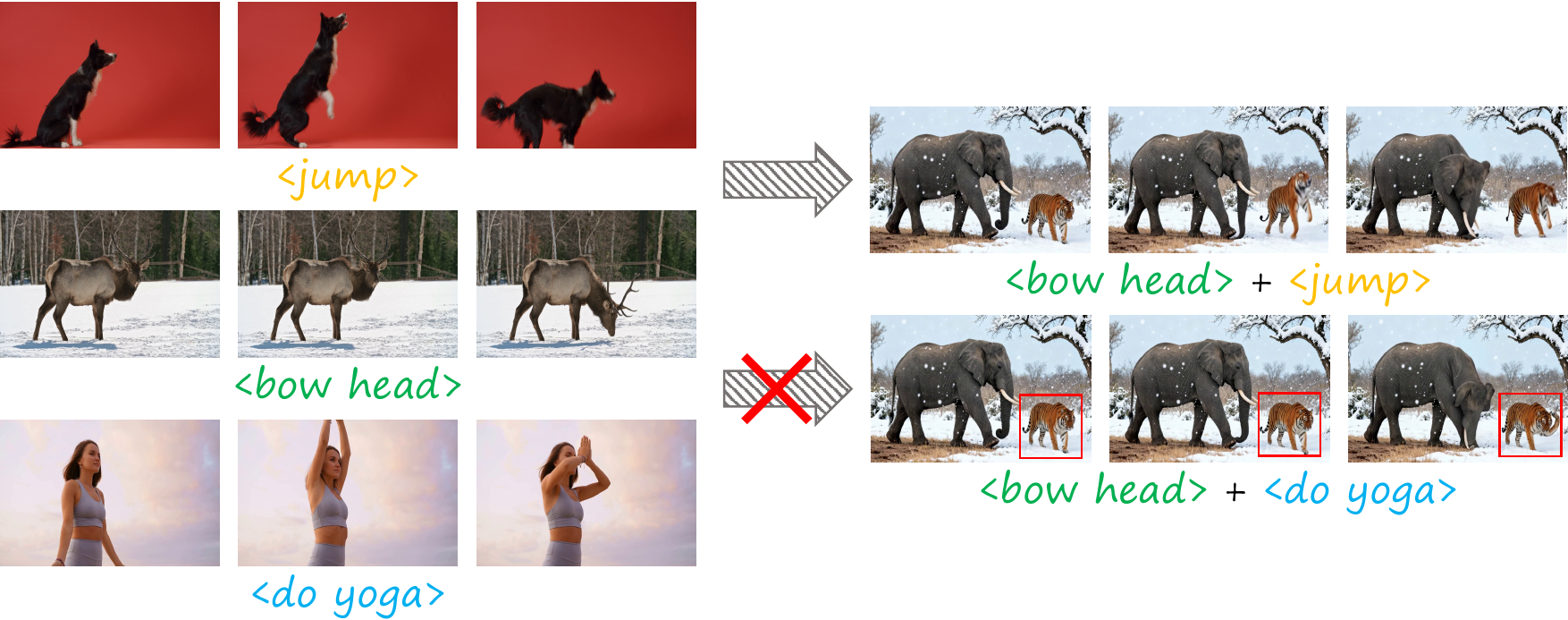}
    \caption{\textbf{Limitation.} Visualization of failure cases in our \Our. The caption for the correctly generated video is: ``An elephant bows its head. A tiger jumps up while barking.'' The caption for the incorrectly generated video is: ``An elephant bows its head. A tiger is doing yoga.''}
    \label{fig:bad_case}
    \vspace{-0.3cm}
\end{figure}
Full text prompts in \cref{fig:additional_res} are:

\begin{itemize}
    \item \textcolor{motionred}{$<$bow head$>$} + \textcolor{motionorange}{$<$exercise$>$} : ``A brown bear bows its head. A character dressed in a Superman costume is exercising with a resistance machine.''

    \item \textcolor{motionorange}{$<$exercise$>$} + \textcolor{motionyellow}{$<$stand up$>$} : ``A man is exercising with a resistance machine. A dog stands up.''
    
    \item \textcolor{motionyellow}{$<$stand up$>$} + \textcolor{motiongreen}{$<$do yoga$>$} : ``A dog stands up. A girl is doing yoga.''
    
    \item \textcolor{motionorange}{$<$exercise$>$} + \textcolor{motionblue}{$<$walk$>$} : ``A girl is exercising with a resistance machine. A dog walks on the street.''
    
    \item \textcolor{motionblue}{$<$walk$>$} + \textcolor{motiondarkblue}{$<$do yoga$>$} : ``A tiger walks on the snow ground with rocks. A woman wearing a light beige blouse and a delicate necklace is doing yoga.''
    
    \item \textcolor{motionpurple}{$<$crouch$>$} + \textcolor{motionorange}{$<$exercise$>$} : ``A female cartoon character in armor is exercising with a resistance machine. A shirtless man wearing black shorts is crouching.''
    
    \item \textcolor{motionpurple}{$<$crouch$>$} + $<$exercise$>$ + \textcolor{motiondarkblue}{$<$do yoga$>$} : ``The left man is crouching. The middle man is performing a fitness exercise. The right man is doing yoga.''
    
    \item $<$exercise$>$ + \textcolor{motiondarkblue}{$<$do yoga$>$} + \textcolor{motionyellow}{$<$stand up$>$} : ``A elderly man with gray hair and a beard is performing a fitness exercise. A man wearing blue suit is doing yoga. A capybara stands up.''
\end{itemize}

\section{Limitation}
Although our method can transfer multiple motions to images with multiple objects, due to the limitation of the model's capabilities, it is difficult to complete the transfer of motion when the structural differences between objects are too large. As shown in \cref{fig:bad_case}, our \Our failed to transfer the human motion \textcolor{motionblue}{$<$do yoga$>$} to the tiger. This deficiency is common in existing models. Therefore, the future goal is to explore better ways to solve the problem of structural adaptability, such as using models with stronger structural compatibility.

\end{document}